\documentclass[11pt,a4paper]{article}
\usepackage{times,latexsym}
\usepackage{url}
\usepackage[T1]{fontenc}

\usepackage[acceptedWithA]{tacl2021v1}

\usepackage{xspace,mfirstuc,tabulary}

\newif\iftaclinstructions
\taclinstructionsfalse %
\iftaclinstructions
\renewcommand{\confidential}{}
\renewcommand{\anonsubtext}{(No author info supplied here, for consistency with
TACL-submission anonymization requirements)}
\newcommand{\instr}
\fi

\iftaclpubformat %

\else

\fi

\usepackage{graphicx}
\usepackage{amsmath}
\usepackage{array}
\usepackage{amsfonts}
\usepackage{booktabs}
\usepackage{subcaption}
\usepackage{stringstrings}
\usepackage{siunitx}
\usepackage{algorithm}
\usepackage{algorithmic}
\usepackage{hyperref}
\usepackage{url}
\usepackage{pifont}
\usepackage{booktabs}
\usepackage{marvosym}
\usepackage{tikz}
\usetikzlibrary{shapes.misc, positioning, fit, backgrounds, arrows.meta}
\usepackage{pgfplots} 
\pgfplotsset{compat=1.17}
\usepackage{colortbl}
\usepackage{caption}
\usepackage{multirow}
\usepackage{inconsolata}
\usepackage{stmaryrd}
\usepackage{xcolor}%
\usepackage{svg}

\definecolor{ourRed}{HTML}{E24A33}
\definecolor{ourBlue}{HTML}{348ABD}
\definecolor{ourPurple}{HTML}{988ED5}
\definecolor{ourGray}{HTML}{777777}
\definecolor{ourLightGray}{HTML}{B8B8B8}
\definecolor{ourYellow}{HTML}{FBC15E}
\definecolor{ourGreen}{HTML}{4D8951}
\definecolor{ourPink}{HTML}{FFB5B8}
\definecolor{oursteelblue}{HTML}{9BB8D7}
\definecolor{ourOrange}{HTML}{FDBA58}
\definecolor{ourWhite}{HTML}{FAFAFA}

\newcommand{\Secref}[1]{Section~\ref{#1}}
\newcommand{\secref}[1]{Section~\ref{#1}}

\newcommand{\Figref}[1]{Figure~\ref{#1}}

\newcommand{\Tabref}[1]{Table~\ref{#1}}
\newcommand{\tabref}[1]{Table~\ref{#1}}

\newcommand{\word}[1]{\emph{#1}}
\newcommand{\lf}[1]{\texttt{#1}}

\newcommand{\phz}{\ifmmode\phantom{0}\else$\phantom{0}$\fi}

\newcommand{\phsz}{\ifmmode\tiny\phantom{0}\else$\tiny\phantom{0}$\fi}

\usepackage[most]{tcolorbox}

\newtcbox{\hlprimarytab}{on line, rounded corners, box align=base, colback=white!10,colframe=white,size=fbox,arc=3pt, before upper=\strut, top=-2pt, bottom=-4pt, left=-2pt, right=-2pt, boxrule=0pt}
\newtcbox{\hlprimarytabg}{on line, rounded corners, box align=base, colback=gray!10,colframe=white,size=fbox,arc=3pt, before upper=\strut, top=-2pt, bottom=-4pt, left=-2pt, right=-2pt, boxrule=0pt}
\newtcbox{\hlsecondarytab}{on line, box align=base, colback=red!10,colframe=white,size=fbox,arc=3pt, before upper=\strut, top=-2pt, bottom=-4pt, left=-2pt, right=-2pt, boxrule=0pt}

\newcommand{\std}[1]{{\tiny\hlprimarytab{{#1}}}}

\usepackage[T1]{fontenc}

\usepackage[utf8]{inputenc}

\usepackage{microtype}

\title{ReCOGS: How Incidental Details of a Logical Form \\ Overshadow an Evaluation of Semantic Interpretation}

\author{%
  \textbf{Zhengxuan Wu}
  \And
  \textbf{Christopher D. Manning}
  \\ Stanford University, USA
  \\ \texttt{\{wuzhengx, manning, cgpotts\}@stanford.edu}
  \And
  \textbf{Christopher Potts}
}

\begin{document}
\maketitle
\begin{abstract}
Compositional generalization benchmarks for semantic parsing seek to assess whether models can accurately compute \emph{meanings} for novel sentences, but operationalize this in terms of \emph{logical form} (LF) prediction. This raises the concern that semantically irrelevant details of the chosen LFs could shape model performance. We argue that this concern is realized for the COGS benchmark \citep{kim2020cogs}. COGS poses generalization splits that appear impossible for present-day models, which could be taken as an indictment of those models. However, we show that the negative results trace to incidental features of COGS LFs. Converting these LFs to semantically equivalent ones and factoring out capabilities unrelated to semantic interpretation, we find that even baseline models get traction. A recent variable-free translation of COGS LFs suggests similar conclusions, but we observe this format is not semantically equivalent; it is incapable of accurately representing some COGS meanings. These findings inform our proposal for ReCOGS, a modified version of COGS that comes closer to assessing the target semantic capabilities while remaining very challenging. Overall, our results reaffirm the importance of compositional generalization and careful benchmark task design.
\end{abstract}

\section{Introduction}

Compositional generalization benchmarks have emerged as powerful tools for diagnosing whether models have learned to systematically interpret natural language \citep{lake2018generalization, kim2020cogs, keysers2019measuring, ruis2020benchmark, wu2021reascan}. The core task is to map sentences to logical forms (LFs), and the goal is to make accurate predictions for held-out examples that include novel grammatical combinations of elements. These tasks are guided by the assumption that natural languages are governed by the  \emph{principle of compositionality}: the meanings of complex phrases are determined by the meanings of their parts and the way those parts combine syntactically, and so the meanings of novel combinations of familiar elements are fully determined as well \cite{Montague74, Partee84, Janssen97}.

The COGS (\textbf{CO}mpositional \textbf{G}eneralization Challenge based on \textbf{S}emantic Interpretation) benchmark of \citet{kim2020cogs} is among the most widely used compositional generalization benchmarks at present, and it is noteworthy for containing assessment splits that almost no present-day models get traction on (\Tabref{tab:cogs-seq2seq-results}). The phenomena are remarkably simple. For instance, in the COGS Object-to-Subject modification split (\Tabref{tab:meta-examples}), models are trained on modified nouns like \word{the cake on the plate} only in grammatical object positions and asked to make predictions about such phrases when they are grammatical subjects. The average score for this split in prior work is roughly 0. Similar patterns obtain for splits requiring generalization to deeper clausal embedding. This looks like strong evidence that present-day models cannot handle basic matters of semantic interpretation.

We argue that this conclusion is hasty. The core issue is one of semantic representation. The goal of COGS is to assess whether models can compute \emph{meanings} compositionally, but this is operationalized as the task of predicting LFs in a specific format. There will in general be innumerable LF formats that express the desired meanings, and we have no reason to privilege any particular one. 

In this paper, we focus on three incidental details of syntactic form that profoundly impact model performance and thus challenge the validity of COGS\@. First, we find that when we make trivial, meaning-preserving modifications to the LFs by removing redundant symbols, we see substantial improvements in model performance (\secref{sec:token-removal}). 

We also identify subtler issues related to variable binding. In COGS, all variables are bound.\footnote
  {That is, the values that each variable can take on are specified by a variable-binding operator such as a quantifier.}
 They appear unbound in the LFs but they are interpreted as existentially closed, as in many theories of dynamic semantics \citep{Kamp81,Heim82,Heim83FCS}. As bound variables, they can be freely renamed with no change to the interpretation as long as the renaming is consistent; $\lf{truck}(x)$ and $\lf{truck}(y)$ are semantically identical in COGS because both variables are implicitly bound with existentials. However, COGS currently requires models to predict the exact identity of variables. This goes well beyond capturing semantics. For neural models that rely on token embeddings, this poses an artificial challenge for test LFs that happen to contain novel variable names 
 or familiar variable names that happen to appear in new contexts. This affects COGS splits involving novel clausal embedding depths (\secref{sec:exp2}) and novel modification patterns (\secref{sec:exp3}). Again, meaning-preserving adjustments to the COGS LFs address these issues and allow even baseline models to succeed.

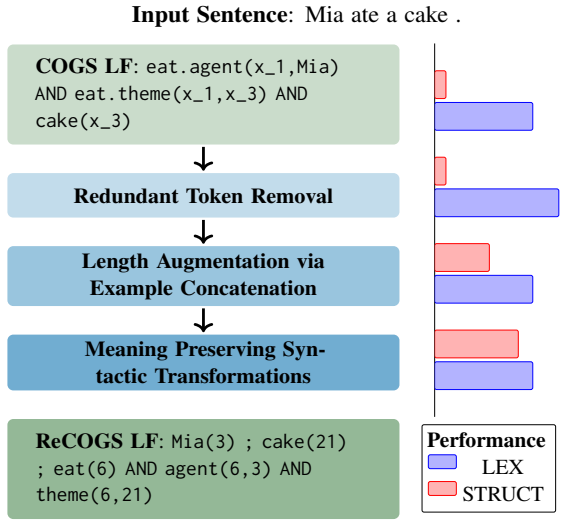
\begin{figure}[t]
  \centering
  \resizebox{1.0\linewidth}{!}{
  \begin{tikzpicture}
  \tikzset{
    every node=[
    draw,
    font=\normalsize,
    shape=rectangle,
    rounded corners=1pt
    ],
    cogs/.style={
      fill=ourGreen!30
    },
    recogs/.style={
      fill=ourGreen!60
    },
    sentence/.style={
        align=center
    },
    transformation1/.style={
        fill=ourBlue!30,
        align=center
    },
    transformation2/.style={
        fill=ourBlue!50,
        align=center
    },
    transformation3/.style={
        fill=ourBlue!70,
        align=center
    },
    ourArrow/.style={
      solid,
      line width=0.5mm,
      ->,
      shorten <=1pt,
      shorten >=1pt,
    }
  };

  \node[sentence, minimum width=10cm, minimum height=0.5cm, text width=10cm, rounded corners=3pt](val0){{\large \textbf{Input Sentence}: Mia ate a cake .}};
  
  \node[cogs, below = 1.4cm of val0.west, anchor=west, minimum width=7cm, minimum height=1.8cm, text width=6cm, rounded corners=3pt](val1){\textbf{COGS LF}: \texttt{eat.agent(x\_1,Mia) AND eat.theme(x\_1,x\_3) AND cake(x\_3)}};
  
  \node[transformation1, below = 0.5cm of val1, minimum width=7cm, text width=5.5cm, minimum height=0.8cm, rounded corners=3pt](val3){\textbf{Redundant Token Removal}};
  \node[transformation2, below = 0.5cm of val3, minimum width=7cm, text width=5cm, minimum height=0.8cm, rounded corners=3pt](val4){\textbf{Length Augmentation via Example Concatenation}};
  \node[transformation3, below = 0.5cm of val4, minimum width=7cm, text width=5cm, minimum height=0.8cm, rounded corners=3pt](val5)
 {\textbf{Meaning Preserving Syntactic Transformations}};
  \node[recogs, below = 0.5cm of val5, minimum width=7cm, minimum height=1.8cm, text width=6cm, rounded corners=3pt](val2){\textbf{ReCOGS LF}: \texttt{Mia(3) ; cake(21) ; eat(6) AND agent(6,3) AND theme(6,21)}};

  \path(val1.south) edge[ourArrow] (val3.north);
  \path(val3.south) edge[ourArrow] (val4.north);
  \path(val4.south) edge[ourArrow] (val5.north);

  \begin{axis}[
    at={(2.5cm,-7.2cm)},
    height = 8.3cm,
    width = 4cm,
    xbar,
    axis y line*=left,
    axis x line       = none,
    tickwidth         = 0pt,
    enlarge y limits  = 0.22,
    enlarge x limits  = 0.1,
    bar width = 0.5cm,
    symbolic y coords = {LaTeX, Tools, Distributions, Editors},
    ytick=\empty,
    yticklabels=\empty,
    legend style={at={(0.4,-0.02)},anchor=north},
    legend image code/.code={
            \draw [#1] (0cm,0cm) rectangle (0.5cm,0.25cm); }
    ]
  \addlegendimage{empty legend}
  \addplot coordinates { (70,LaTeX)         (70,Tools)
                         (88,Distributions)  (70,Editors) };
  \addplot coordinates { (60,LaTeX)         (40,Tools)
                         (10,Distributions)   (10,Editors)  };
  
   \addlegendentry{\hspace{-.6cm}\textbf{Performance}}
   \addlegendentry{LEX}
   \addlegendentry{STRUCT}
  \end{axis}

\end{tikzpicture}}
  \caption{Converting COGS LFs into semantically equivalent LFs greatly impacts model performance: removing redundant tokens increases performance on the lexical (LEX) tasks, while length augmentation and meaning-preserving syntactic transformations help on the harder structural (STRUCT) tasks. ReCOGS incorporates these lessons while also decoupling variable names from linear position. The result is a more purely semantic task that remains extremely challenging for present-day models.
  }
  \label{fig:main}
\end{figure}

\begin{table*}
\centering
\resizebox{1.0\linewidth}{!}{
\begin{tabular}{l *{5}{c} }
\toprule
& \multicolumn{3}{c}{\textbf{STRUCT}} & \textbf{LEX} & \textbf{Overall} \\
\textbf{Model} & Obj PP $\rightarrow$ Subj PP & CP Recursion & PP Recursion &  & \% \\
\midrule
BART~\cite{lewis2019bart} & 0 & {\phz}0 & 12 & 91 & 79\rlap{$^\dagger$} \\
BART+syn~\cite{lewis2019bart} & 0 & {\phz}5 & {\phz}8 & 80  & 80\rlap{$^\dagger$} \\
T5~\cite{raffel2020exploring} & 0 & {\phz}0 & {\phz}9 & 97 & 83\rlap{$^\dagger$} \\ \midrule
\citealt{kim2020cogs} & 0 & {\phz}0 & {\phz}0 & 73 & 63 \\
\citealt{ontanon2021making} & 0 & {\phz}0 & {\phz}0 & 53 & 48 \\
\citealt{akyurek2021lexicon} & 0 & {\phz}0 & {\phz}1 & 96 & 82 \\
\citealt{conklin2021meta} & 0 & {\phz}0 & {\phz}0 & 88 & 75 \\
\citealt{csordas2021devil} & 0 & {\phz}0 & {\phz}0 & 95 & 81 \\ \midrule
\citealt{zheng2021disentangled} & 0 & 25 & 35 & 99 & 88\rlap{$^\ddagger$} \\
\bottomrule
\end{tabular}}
\caption{Results on the COGS benchmark for different generalization splits, including recent seq2seq models specialized for COGS. {}$^\dagger$Models use pretrained weights, and their results are copied from \citet{yao2022structural}. {}$^\ddagger$Model uses pretrained weights and is hyperparameter tuned using data sampled from the generalization splits. Our focus is on the factors behind the strikingly bad performance of all models, but especially the models that are not pretrained, on the structural generalization splits.} \label{tab:cogs-seq2seq-results}
\end{table*}

\begin{table*}[t!]
    \centering

    \resizebox{1.0\linewidth}{!}{

    \begin{tabular}{p{0.19\linewidth} @{\hspace{0.2cm}} c @{\hspace{0.3cm}} p{0.76\linewidth}}
    \toprule
      \textbf{Case}  & \textbf{Split} & \textbf{Example}  \\
      \midrule
      Subj $\rightarrow$ Obj Proper & \emph{Train} & \textbf{Lina} gave the bottle to John .  \\
      & (LF) & \lf{*bottle(x\_3) ; give.agent(x\_1,Lina) AND give.theme(x\_1,x\_3) AND give.recipient(x\_1,John)}  \\
      \cmidrule(lr){2-3}
      & \emph{Gen.} & A cat rolled \textbf{Lina} . \\
      & (LF) & \lf{cat(x\_1) AND roll.agent (x\_2,x\_1) AND roll.theme(x\_2,Lina)}  \\
      
      \midrule
      Prim $\rightarrow$ Subj Proper & \emph{Train} & \textbf{Paula}  \\
      & (LF) & \lf{Paula}  \\
      \cmidrule(lr){2-3}
      & \emph{Gen.} & \textbf{Paula} painted a cake . \\
      & (LF) & \lf{paint.agent(x\_1,Paula) AND paint.theme(x\_1,x\_3) AND cake(x\_3)}  \\

      \midrule
      Prim $\rightarrow$ Obj Proper & \emph{Train} & \textbf{Paula}  \\
      & (LF) & \lf{Paula}  \\
      \cmidrule(lr){2-3}
      & \emph{Gen.} & James rolled \textbf{Paula} . \\
      & (LF) & \lf{agent(x\_1,James) AND roll.theme(x\_1,Paula)}  \\

      \midrule
      Obj PP $\rightarrow$ Subj PP & \emph{Train} & Emma ate \textbf{the cake on the table} . \\
      & (LF) & \lf{*cake(x\_3) ; *table(x\_6) ; eat.agent(x\_1,Emma) AND eat.theme( x\_1,x\_3) AND cake.nmod.on(x\_3,x\_6)}  \\
      \cmidrule(lr){2-3}
      & \emph{Gen.} & \textbf{The cake on the table} burned . \\
      & (LF) & \lf{*cake(x\_1) ; *table(x\_4) ; cake.nmod.on(x\_1,x\_4) AND burn.theme( x\_5,x\_1)}  \\

      \midrule
      CP Recursion & \emph{Train} & Noah knew \textbf{that} Emma said \textbf{that} the cat painted . \\
      & (LF) & \lf{*cat(x\_7) ; know.agent(x\_1,Noah) AND know.ccomp(x\_1,Emma) AND say.agent(x\_4,Emma) AND say.ccomp(x\_4,x\_7) AND paint.agent( x\_8,x\_7)}  \\
      \cmidrule(lr){2-3}
      & \emph{Gen.} & Noah knew \textbf{that} Emma said \textbf{that} John saw \textbf{that} the cat painted . \\
      & (LF) & \lf{*cat(x\_10) ; know.agent(x\_1,Noah) AND know.ccomp(x\_1,Emma) AND say.agent(x\_4,Emma) AND say.ccomp(x\_4,x\_7) AND see.agent( x\_7,John) AND see.ccomp(x\_7,x\_10) AND paint.agent(x\_11,x\_10)}  \\

      \midrule
      PP Recursion & \emph{Train} & John saw the ball \textbf{in} the bottle \textbf{in} the box . \\
      & (LF) & \lf{*ball(x\_3) ; *bottle(x\_6) ; *box(x\_9) ; see.agent(x\_1,John) AND see.theme(x\_1,x\_3) AND ball.nmod.in(x\_3,x\_6) AND bottle .nmod.in(x\_6,x\_9)} \\
      \cmidrule(lr){2-3}
      & \emph{Gen.} & John saw the ball \textbf{in} the bottle \textbf{in} the box \textbf{on} the floor . \\
      & (LF) & \lf{*ball(x\_3) ; *bottle(x\_6) ; *box(x\_9) ; *floor(x\_12) ; see.agent( x\_1,John) AND see.theme(x\_1,x\_3) AND ball.nmod.in(x\_3,x\_6) AND bottle.nmod.in(x\_6,x\_9) AND box.nmod.on(x\_9,x\_12)} \\
      
      \bottomrule
    \end{tabular}}
    \caption{Representative COGS generalization (\emph{Gen.})\ splits with logical forms (LFs). Due to space constraints, we use simplified versions of examples included in the dataset. LFs are detokenized for readability; tokenized examples can be found in \Tabref{tab:examples}.}
    \label{tab:meta-examples}
\end{table*}

At the same time, we emphasize that COGS does already contain instances of variable binding relationships that are challenging for all present-day models. A recent proposal by \citet{qiu2021improving} to map all COGS LFs to variable-free forms is not meaning preserving precisely because it cannot handle these binding relationships properly (\secref{sec:related}). Thus, the high scores models have obtained for this COGS variant are partly illusory.

We close with a proposal for a revised version of COGS, ReCOGS, that incorporates the above insights (\secref{sec:recogs}). ReCOGS is easier than COGS in some respects and harder in others. Through ablation studies in \Secref{sec:recogs}, we show that the original linear variable binding of COGS actually make some aspects of COGS artificially easier, and can prevent us from accurately accessing compositional generalization of models. As we noted above, any particular choice of LFs will be somewhat arbitrary relative to our goals, but we feel that ReCOGS comes closer to assessing whether our models possess the compositional generalization capabilities that \citet{kim2020cogs} identified as essential, and our findings suggest that present-day models continue to struggle in these areas.\footnote{We release ReCOGS and our associated experiment code at \url{https://github.com/frankaging/ReCOGS}.}

\section{Background: COGS Benchmark}

COGS consists of input--output pairs mapping English sentences to LFs. The dataset is generated using a rule-based approach, which allows COGS to maintain systematic gaps between training and different evaluation splits. COGS LFs are based on a Neo-Davidsonian view of verbal arguments~\cite{parsons1990events}, in which verbs introduce event arguments and participants are linked to those events via thematic role predicates.
As discussed in the original paper~\cite{kim2020cogs}, the LFs are created by post-processing the simplified ones defined in \citet{reddy2017universal}. The LFs are purely conjunctive (conjunction is denoted by \lf{;} and \lf{AND}), and conjuncts are sorted by their variable names, which are determined by the position of the head phrase in the sentence (see \Tabref{tab:meta-examples} for examples). Event predicates for nominals are not included. Definite and indefinite phrases are formally distinguished. All COGS variables are bound; the definiteness operator \lf{*} binds variables locally to its conjunct, and all other variables are interpreted as bound by implicit widest-scope existential quantifiers.

By convention, the subscripts on variables in COGS correspond to the 0-indexed position of the corresponding word in the input sentence. Thus, the LF for \word{A cat rolled Lina} includes a conjunct \lf{cat(x\_1)}, which indicates \word{cat} is the first word in the sentence, whereas the LF for \word{Lina rolled a cat} includes a conjunct \lf{cat(x\_3)}, which indicates \word{cat} is the third word in the sentence.

The COGS evaluation metric is percent exact string identity of logical forms. COGS provides a single training split as well as standard in-distribution (IID) validation splits. In addition, the dataset includes \emph{generalization splits} designed around types of examples that were not seen in training but seem to be natural extrapolations of examples seen in training under assumptions of compositionality, as exemplified in \Tabref{tab:meta-examples}. The generalization splits cover five scenarios:
\begin{enumerate}\setlength{\itemsep}{0pt}
\item\label{split:1} Interpreting novel pairings of primitives and grammatical roles (e.g., Subj $\rightarrow$ Obj Proper).

\item Verb argument structure alternation (e.g., Active $\rightarrow$ Passive).

\item Sensitivity to verb class (e.g., Agent NP $\rightarrow$ Unaccusative subject).

\item Interpreting novel combinations of modified phrases and grammatical roles (e.g., Object PP $\rightarrow$ Subject PP).

\item Generalizing phrase nesting to unseen depths (e.g., CP Recursion). 
\end{enumerate}
The first three fall under lexical generalization, and the final two require structural generalization. 

It is noteworthy that, for the splits in group~\ref{split:1}, there is only a single training example for the primitive with a single input and output token (e.g., ``Paula'' $\rightarrow$ \texttt{Paula}). This leads to higher variance across these related splits; we suspect that model performance for these cases is affected by when this single example is seen by the model during training.

\begin{table*}[ht]
    \centering
    \resizebox{1.0\linewidth}{!}{
    \begin{tabular}{p{0.35\linewidth} p{0.65\linewidth}}
    \toprule
      \textbf{Variant (Token Removal Set)}  & \textbf{Logical Form} \\ \midrule
      COGS & \lf{* boy ( x \_ 3 ) ; hold . agent ( x \_ 1 , Liam ) AND hold . theme ( x \_ 1 , x \_ 3 ) AND boy . nmod . beside ( x \_ 3 , x \_ 6 ) AND table ( x \_ 6 )} \\
      Token Removal (\{\lf{x} \lf{\_}\}) & \lf{* boy ( 3 ) ; hold . agent ( 1 , Liam ) AND hold . theme ( 1 , 3 ) AND boy . nmod . beside ( 3 , 6 ) AND table ( 6 )} \\
      Token Removal (\{\lf{x} \lf{\_} \lf{(} \lf{)}\}) & \lf{* boy 3; hold . agent 1 , Liam AND hold . theme 1 , 3 AND boy . nmod . beside 3 , 6 AND table 6} \\
      Token Removal (\{\lf{x} \lf{\_} \lf{(} \lf{)} \lf{,}\}) & \lf{* boy 3; hold . agent 1 Liam AND hold . theme 1 3 AND boy . nmod . beside 3 6 AND table 6} \\
      \bottomrule
    \end{tabular}}
    \caption{Removing redundant tokens from the LF for the sentence \word{Liam held the boy beside a table .}}
    \label{tab:examples}
\end{table*}

\section{Related Work}\label{sec:related}

\paragraph{Approaches to COGS}
Researchers have adopted a variety of approaches to solving COGS, including grammar-based rules \cite{herzig2020span}, lexicons or lexicon-style alignments incorporated into seq2seq models \cite{akyurek2021lexicon, zheng2020compositional}, modified Transformer models for better-structured representations \cite{oren2020improving, zheng2021disentangled, bergen2021systematic, csordas2021devil}, meta-learning \cite{conklin2021meta}, tree-like neural parsers \cite{weissenhorn2022compositional}, a grammar-enhanced seq2seq learner~\cite{wang2022hierarchical}, and various data augmentation techniques \cite{qiu2021improving}. 

\paragraph{COGS Artifacts}
A number of recent papers investigate artifacts in compositional generalization benchmarks such as SCAN~\cite{patel2022revisiting} and ReaSCAN~\cite{sikarwar2022can}. \citet{csordas2021devil} focus on COGS\@. They examine potential pitfalls that might lead us to underestimate a model's ability to generalize. By carefully exploring the effects of relative positional embeddings and training time, they are able to increase overall model performance from 35\% to 81\% on the generalization splits.
Our work complements these findings by focusing on issues of semantic representation. 

\paragraph{A Flawed Variable-Free COGS Representation}
\citet{qiu2021improving} propose a variable-free version of COGS and show that models score much better on the format (which is used by \citet{drozdov2022compositional} in experiments with GPT-3). However, the representations of \citeauthor{qiu2021improving}\ do not preserve the meaning of COGS examples. COGS embeds variable-binding challenges that the variable-free forms artificially side-step. For instance, their formalism represents \word{A zebra needs to walk} as 
\\[1ex]
\resizebox{1\linewidth}{!}{%
    \lf{need(agent=zebra,xcomp=walk(agent=zebra))}%
}
\\[1ex]
which means `a zebra needs a zebra to walk', with two unlinked occurrences of the indefinite \word{a zebra}. Consider a model where zebra $a$ needs zebra $b$ to walk, for $a \neq b$, and no zebra $a$ is such that $a$ needs $a$ to walk. The above LF is true in this model but our original sentence is not \citep{mccawley1968lexical}. In addition, these variable-free forms would be unable to represent quantifier-binding relationships like \word{Every zebra ate its meal} as well as simple reflexives like \word{A zebra saw itself}. Since variable binding arguably reflects one of the deepest challenges of natural language interpretation, we argue that these phenomena should not be marginalized.\footnote{The results of \citet{curry1958combinatory} ensure that there is some variable-free representation scheme that can capture these phenomena. It may be worthwhile to develop and explore such representations as alternative LFs for COGS.}

\section{Experiments}

We report on experiments studying semantic representations for COGS\@. For any modifications we apply to the dataset, we ensure there is no data leakage, as such leakage trivializes the generalization tests (as evidenced by the consistently very high results for COGS IID test splits in prior work).

\subsection{Methods}

\paragraph{Architectures}
Following the original COGS paper~\cite{kim2020cogs}, we train encoder--decoder models with two model architectures: LSTMs~\cite{hochreiter1997long} and Transformers~\cite{vaswani2017attention}. We adopt similar configurations to the original paper. For the LSTMs, we use a 2-layer LSTM as the encoder with global attention and a dot-product scoring function, and a 2-layer LSTM as the decoder with a hidden dimension size of 512. For the Transformers, we use two 2-layer Transformer blocks with 4 attention heads and learnable absolute positional embeddings\footnote{We found that using relative embeddings or initializing embeddings with periodic functions did not improve performance, corroborating findings of \citet{zheng2021disentangled}.} with a hidden dimension size of 300. 

The LSTM architecture has approximately 9M parameters whereas the Transformer architecture has approximately 4M parameters.

\paragraph{Training Details}
We use cross-entropy loss and a fixed number of training epochs (200 for LSTMs and 300 for Transformers), as previous work suggests that early stopping hurts model performance~\cite{csordas2021devil}. We set the batch size to 128 for the Transformers and 512 for the LSTMs. We train with a fixed learning rate of 8$\times$10$^{-4}$ for the LSTMs, and 1$\times$10$^{-4}$ for the Transformers. 

We use a single NVIDIA GeForce RTX 3090 24GB GPU to train our models. For LSTMs, the training time ranges from 0.5 hours to 5 hours. For Transformers, it ranges from 0.3 hours to 3 hours. We run each experiment with 20 distinct random seeds. Additional training details
can be found in our code repository.

\paragraph{No Pretraining}
We train all of our models from scratch, without any pretraining, to ensure that we are not introducing outside information that could be relevant to the COGS generalization tasks \citep[q.v.][]{kim2022uncontrolled}.

\newcommand{\gradient}[2]{
    \ifdimcomp{#1pt}{>}{\maxval pt}{#1}{
    \ifdimcomp{#1pt}{<}{\minval pt}{#1}{
         \pgfmathparse{min(int(round( ((abs(#1 - #2))) * 3 )), 100)}
        \xdef\tempa{\pgfmathresult}
        \colorbox{high!\tempa!low!\opacity}{{ }#1{ }}
    }}
 }

\newcommand{\gradientdeep}[2]{
    \ifdimcomp{#1pt}{>}{\maxval pt}{#1}{
    \ifdimcomp{#1pt}{<}{\minval pt}{#1}{
         \pgfmathparse{min(int(round( ((abs(#1 - #2))) * 5 )), 100)}
        \xdef\tempa{\pgfmathresult}
        \colorbox{high!\tempa!low!\opacity}{{ }#1{ }}
    }}
 }
 
\begin{table*}[ht!]
\centering
\resizebox{1.0\linewidth}{!}{
\begin{tabular}{l >{\centering\arraybackslash}p{2.5cm} >{\centering\arraybackslash}p{2.5cm} >{\centering\arraybackslash}p{2.5cm} >{\centering\arraybackslash}p{2.5cm} }
\toprule
& \multicolumn{3}{c}{\textbf{LEX}} & \\
\textbf{Model (Token Removal Set)} & Subj $\rightarrow$ Obj Proper & Prim $\rightarrow$ Obj Proper & Prim $\rightarrow$ Subj Proper & \textbf{Overall} \\
\midrule
LSTM & 5.3\std{[1.4, 9.3]} & {\phsz}21.9\std{[10.9, 32.9]} & 69.0\std{[51.6, 86.3]} & 32.1\std{[28.2, 36.0]} \\
+ Tokens Removal (\{\lf{x} \lf{\_}\}) & {\phsz}5.1\std{[0.2, 10.0]} & 18.1\std{[8.7, 27.4]} & 76.5\std{[60.9, 92.0]} & 34.9\std{[31.5, 38.2]} \\
+ Tokens Removal (\{\lf{x} \lf{\_} \lf{(} \lf{)}\}) & {\phsz}7.2\std{[1.6, 12.9]} & {\phsz}24.6\std{[14.5, 34.8]} & 88.1\std{[78.2, 98.1]} & 39.6\std{[36.7, 42.5]} \\
+ Tokens Removal (\{\lf{x} \lf{\_} \lf{(} \lf{)} \lf{,}\}) & 5.5\std{[1.2, 9.8]} & 14.7\std{[5.8, 23.6]} & 65.4\std{[45.6, 85.3]} & 36.5\std{[32.8, 40.2]} \\
\midrule
Transformer & 74.4\std{[71.9, 76.9]} & {\phsz}62.4\std{[60.1, 64.6]} & 97.6\std{[96.4, 98.8]} & 81.3\std{[80.7, 81.9]} \\
+ Tokens Removal (\{\lf{x} \lf{\_}\}) & 91.9\std{[90.0, 93.9]} & {\phsz}81.1\std{[74.5, 87.7]} & 99.2\std{[98.6, 99.7]} & 83.6\std{[82.9, 84.3]} \\
+ Tokens Removal (\{\lf{x} \lf{\_} \lf{(} \lf{)}\}) & 86.0\std{[83.2, 88.9]} & {\phsz}71.2\std{[61.3, 81.1]} & {\phsz}97.8\std{[95.5, 100.2]} & 82.8\std{[82.1, 83.4]} \\
+ Tokens Removal (\{\lf{x} \lf{\_} \lf{(} \lf{)} \lf{,}\}) & 88.3\std{[84.4, 92.2]} & {\phsz}73.4\std{[65.8, 81.0]} & 99.0\std{[98.7, 99.4]} & 83.4\std{[82.8, 84.0]} \\
\bottomrule
\end{tabular}}
\caption{Results on the COGS lexical generalization splits for different token removal scheme. We report means (over 20 evaluations) with bootstrapped 95\% confidence intervals.}
\label{tab:cogs-results-token}
\end{table*}

\begin{figure}[tp]
  \centering
  \includesvg[width=1.0\columnwidth]{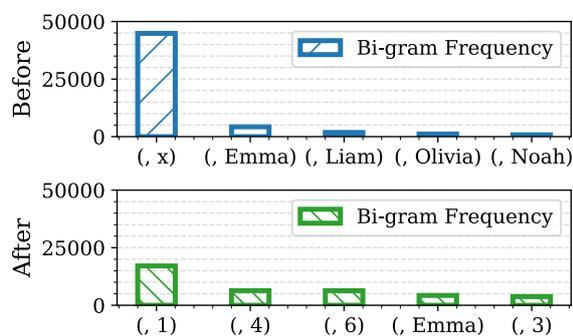}
  \caption{The frequencies of bigrams in the training data starting with \lf{,} become more balanced  after removing two incidental tokens \{\lf{x} \lf{\_}\}.}
  \label{fig:bigrams}
\end{figure}

\subsection{Experiment 1: Removing Redundant Tokens from LFs}\label{sec:token-removal}

Our first experiments involve only very trivial modifications to COGS LFs: we remove some redundant tokens from the LFs, with no other modifications to the examples, and we study the effects this has on the lexical generalization splits in COGS. (These modifications alone do not significantly improve results on the structural generalization splits.)

\paragraph{LF Modifications}
\Tabref{tab:examples} summarizes our redundant token removal strategies. The first row gives a COGS example. Each noun or verb is associated with a variable representing its token position in the input sentence. Concretely, these variables are given as three whitespace-separated elements: $\lf{x \_\ N}$, where \lf{N} is a numeral. The initial \lf{x}, the underscore, and the spaces do not contribute to variable identity in any way. Thus, we can remove these prefixes without changing the semantics of the LFs. The second row of \tabref{tab:examples} depicts removal of the underscore, and the third row depicts removal of the entire prefix. Additionally, we experiment with different token removal schemes by removing redundant punctuation in the set \{\lf{,}\lf{(} \lf{)}\}.

\paragraph{Results}
\Tabref{tab:cogs-results-token} illustrates model performance on the three most demanding lexical generalization splits when models are trained and assessed with these minor variants of COGS LFs. Our primary finding is that model performance is highly sensitive to redundant token removal. By removing the prefix \lf{x  \_}, the Transformer achieves nearly 30\% better performance on average for the Primitive $\rightarrow$ Object Proper Noun split, and the LSTM's performance improves 27\% for the Primitive $\rightarrow$ Subject Proper Noun split after removing the prefix \lf{x  \_} and parenthesis tokens. 

Our findings show that model performance can vary substantially across different semantically equivalent LFs. In addition, while model performance is stable in the Overall evaluation, some of the splits show a high degree of sensitivity to the random seed. Nonetheless, even in these splits, the performance increases that stem from redundant token removal are consistently large enough to be robust to this variance.

\begin{figure*}[tp]
\begin{subfigure}{0.48\textwidth}
  \centering
  \includesvg[width=1.0\columnwidth]{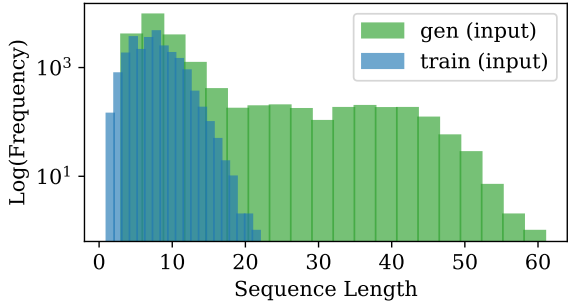}
  \caption{Input sentences}
  \label{fig:length_input}
\end{subfigure}
\hfill
\begin{subfigure}{0.48\textwidth}
  \centering
  \includesvg[width=1.0\columnwidth]{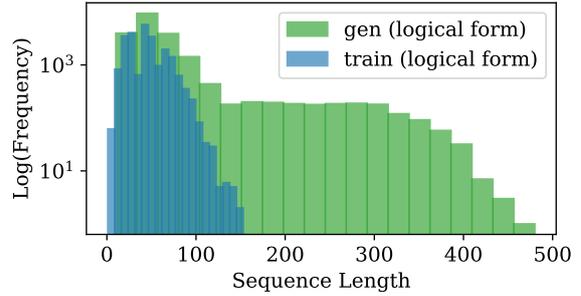}
  \caption{Output logical forms}
  \label{fig:length_lf}
\end{subfigure}
\caption{Sequence length distributions for the COGS training split and the generalization splits. The generalization split has inputs and logical forms with lengths completely unseen in the training set.}
\label{fig:lengths}
\end{figure*}

We hypothesize that the improvements that come from removing these redundant tokens 
derive from simple considerations of how sequence models operate. As shown in \Figref{fig:bigrams}, the bigram \lf{, x} appears 44,846 times, whereas the bigram \lf{, Emma} appears 4,279 times (\lf{Emma} is the most frequent proper noun in the training split). Thus, the conditional bigram probability $P( \lf{x} \mid \lf{,})$ is significantly larger than $P(\lf{Emma} \mid \texttt{,})$. Removing the prefix balances the dataset in this respect because we then mostly care about $P( \lf{N} \mid \lf{,})$ for different values of \lf{N}\@. This helps the decoder to generate less skewed label distributions. This corroborates the findings of \citet{yao2022structural}, who show that the decoder is heavily biased towards generating seen n-grams.

\begin{figure}[tp]
  \centering
  \includesvg[width=1.0\columnwidth]{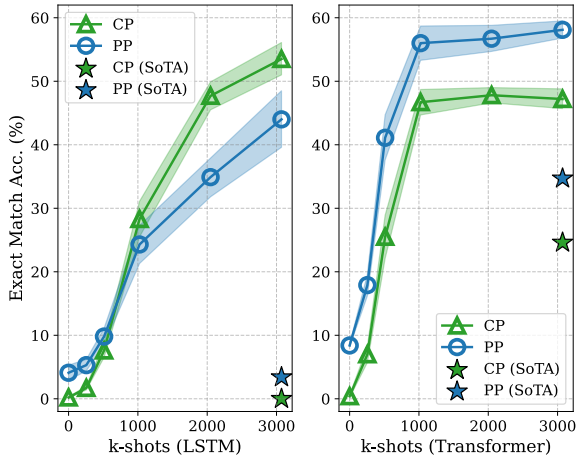}
  \caption{Adding $k$ items with concatenated training examples to give exposure to long sequences greatly improves structural generalization on COGS for both LSTM-based and transformer-based models. (Transformer-based) SoTA performance is taken from \citet{zheng2020compositional}. The plots show means (of 20 runs) with 95\% confidence interval.}
  \label{fig:recursive_gen_plot}
\end{figure}

\subsection{Experiment 2: Separating Structural and Length Generalization} \label{sec:exp2}

The splits that focus on recursive phrase embedding are among the hardest problems in COGS. When training, models see multiply embedded examples like [\textit{John knows that} [\textit{Noah knows that \ldots}]$_{\text{CP}}$]$_{\text{CP}}$ containing nestings of depth 0--2. At test time, they see examples involving strictly greater depth (3--12). COGS includes two constructions that allow nesting: sentential complements (nested CPs) and nominal PP modifiers (nested PPs). 

The nature of these COGS splits creates a strict relationship between recursion depth and sequence length: deeper recursion leads to longer sequences. \Figref{fig:lengths} summarizes the situation when it comes to input sentences and output logical forms. As expected, we see non-overlapping long-tail distributions over longer sequences. 

This is perfectly well-posed as a generalization task simultaneously assessing models on both longer lengths and deeper recursion. However, COGS binds these two tasks together in a way that takes us outside of the goal of assessing semantic generalization. Recall that COGS variables are named according to their linear position. This virtually guarantees that there will be variables whose numerical components are encountered only at test time. For instance, if \lf{47} is the largest variable index seen at train time, then \lf{48} and above will be encountered only in testing. For models that rely on embeddings, these new variables will have random representations at test time. In other words, COGS is underspecified in a way that prevents models from learning novel positional embeddings or embeddings for novel positional indices. However, there is nothing privileged about this particular variable naming scheme; as we discussed earlier, bound variables can be freely renamed as long as this does not change any binding relationships.

Relatedly, for models like Transformers with token positional embeddings, some of the position representations will be encountered only at test time. It is fair to say that this is a limitation of these models that COGS is exposing, but it also further reinforces the concern that length generalization may be totally overshadowing the challenge of generalizing to novel recursion depths.

\paragraph{LF Modifications}
To overcome the challenge of length generalization, we augment the training data by concatenating existing examples together, reindexing the output LFs with higher positional indexes, and gluing them together as conjunctive terms. In doing this, we do not add any semantically new claims to the COGS training set, and we do not create any new recursive structures. We simply ensure that the relevant token and position indices are not random at test time. Similar concatenation experiments have been shown effective in understanding model artifacts for language model acceptability tasks \cite{sinha2022language}.

\paragraph{Results}
\Figref{fig:recursive_gen_plot} presents our results. 
We gradually introduce $k$ augmented examples into our training set, where $k \in \{256, 512, 1024, 2048, 3072\}$. %
Our results indicate that the previously seen catastrophic failure of structural generalization over nested clauses is largely due to the fact that models are not trained with longer sequences. When the sequence length issue is addressed, the models appear to be very capable at handling novel recursion depths. Indeed, our models now far surpass published state-of-the-art results on these tasks.

We further note that the failures are not due to our setup of relative positional embeddings, as fine-tuning a \texttt{T5-base} model with fixed window-based relative positional embeddings remains 0.0\% on both splits, as shown in \Tabref{tab:cogs-seq2seq-results}.

\begin{table*}[ht]
    \centering
    \resizebox{1.0\linewidth}{!}{
    \begin{tabular}{p{0.15\linewidth} p{0.33\linewidth} p{0.51\linewidth}}
    \toprule
      \textbf{Variant}  & \textbf{Sentence}  & \textbf{Logical Form (LF)} \\ \midrule
      Preposing \newline + Filler Words & \textbf{The box in the tent} Emma was \textbf{um um} lended . & \texttt{*box(x\_1) ; *tent(x\_4) ; box.nmod.in(x\_1 ,x\_4) AND lend.theme(x\_7,x\_1) AND lend .recipient( x\_7,Emma)} \\
      Participial Verb Phrase (\textit{Subj}) & A leaf \textbf{painting the spaceship} froze . & \texttt{*spaceship(x\_4) ; leaf(x\_1) AND leaf.acl .paint(x\_1,x\_4) AND freeze.theme(x\_5,x\_1)} \\
      \bottomrule
    \end{tabular}}
    \caption{Modifications of COGS input and output sequences that we use to diagnose artifacts in the original semantic representation. LFs are detokenized for readability; cf.\ \Tabref{tab:examples} for tokenized examples.}
    \label{tab:examples2}
\end{table*}

\subsection{Experiment 3: Variable Name Binding Prevents Generalization}\label{sec:exp3}

The hardest COGS split based on published numbers seems to be the structural generalization task that involves interpreting novel combinations of modified phrases and grammatical role -- e.g., interpreting subject noun phrases with PP modifiers when the train set includes only object noun phrases with such modifiers (\word{Noah ate \textbf{the cake on the plate}.}\ $\rightarrow$ \word{\textbf{The cake on the plate} burned}). To the best of our knowledge, all prior seq2seq models have completely failed to get traction here (\tabref{tab:cogs-seq2seq-results}). Our goal in this section is to understand more deeply why this split has proven recalcitrant.

To start, we observe that this split is arguably different conceptually from the others. The train set contains only object-modifying PPs. It is quite reasonable for a learner to infer from this situation that PPs are allowed only in this position. Natural languages manifest a wide range of subject/object asymmetries, and learners presumably induce these at least in part from an absence of certain kinds of inputs in their experience. Thus, there is a case to be made that this split is not strictly speaking \emph{fair} in the sense of \citet{geiger2019posing}: we have a generalization target in mind as analysts, but this target is not uniquely defined by the available data in a way that would invariably lead even an ideal learner to the desired conclusion.

That said, we feel it is reasonable to explore whether models can learn the sort of theory that would naturally allow them to make correct predictions about these particular novel inputs, even if their training experiences run counter to that. However, COGS erects another obstacle to this goal by numbering variables using a word's position index in the sentence. At train time, the model sees only PP modifiers like the one in the first row of \Tabref{tab:examples}. This associates PP modifiers with a particular range of tokens (variable numerals). As before, this indexing scheme is not a semantic matter, but rather a superficial matter of representation. However, it can also be seen as very direct supervision about the limited distribution of these modifiers: they can associate only with relatively high variable indices.

Additionally, for models with learned positional indices, all modifier phrase tokens are associated with a particular set of relatively high positional index values. These models thus further reinforce the distributional inference that such phrases cannot appear in subjects.

Both of the above concerns are supported by error analyses on vanilla models trained on the original COGS LFs. For instance, to correctly generate the LF for subject-modifying PPs, the model has to predict the LF term for a PP modifier as the first conjunctive clause after the semicolon (e.g, $\lf{cake .nmod .on}$). However, 0\% of the model's predictions follow this pattern, as this never happens during training. Likewise, since no subject-modifying PPs are seen in training, the variables mentioned in  PP clauses are never associated with the subject. 
The effect of these patterns on model behavior is clear: the first variable of any generated modified phrase clause always refers to a token position seen during training.

To isolate these effects from the structural generalization, we propose three data modification strategies. None of these strategies change the set of meanings expressed by COGS. Rather, they make adjustments to the syntax that, according to the COGS indexing rules, automatically expand the range of variables and positional indices associated with modifier phrases. Examples for each strategy are given in \tabref{tab:examples2}.

\newcommand{\gradientsecond}[2]{
    \ifdimcomp{#1pt}{>}{\maxval pt}{#1}{
    \ifdimcomp{#1pt}{<}{\minval pt}{#1}{
         \pgfmathparse{min(int(round( ((abs(#1 - #2))) * 1.2 )), 100)}
        \xdef\tempa{\pgfmathresult}
        \colorbox{high!\tempa!low!\opacity}{{ }#1{ }}
    }}
 }

\newcommand{\gradientseconddeep}[2]{
    \ifdimcomp{#1pt}{>}{\maxval pt}{#1}{
    \ifdimcomp{#1pt}{<}{\minval pt}{#1}{
         \pgfmathparse{min(int(round( ((abs(#1 - #2))) * 5 )), 100)}
        \xdef\tempa{\pgfmathresult}
        \colorbox{high!\tempa!low!\opacity}{{ }#1{ }}
    }}
 }

\begin{table*}[ht!]
\centering
\resizebox{1.0\linewidth}{!}{
\begin{tabular}{l >{\centering\arraybackslash}p{2.5cm} >{\centering\arraybackslash}p{2.5cm} >{\centering\arraybackslash}p{2.5cm} >{\centering\arraybackslash}p{2.5cm} }
\toprule
& \multicolumn{3}{c}{\textbf{STRUCT}} & \\
\textbf{Model} & Obj PP $\rightarrow$ Subj PP & CP Recursion & PP Recursion & \textbf{Overall} \\
\midrule
LSTM & 0.0\std{[0.0, 0.0]} & 0.2\std{[0.1, 0.3]} & {\phsz}4.1\std{[2.9, 5.3]} & 32.1\std{[28.2, 36.0]} \\
+ Preposing & 8.1\std{[6.6, 9.5]} & 0.3\std{[0.1, 0.4]} & {\phsz}3.4\std{[2.1, 4.6]} & 32.7\std{[28.6, 36.7]} \\
+ Preposing + Filler Words & {\phsz}8.7\std{[7.1, 10.2]} & 0.8\std{[0.6, 1.0]} & {\phsz}0.1\std{[0.0, 0.2]} & 33.9\std{[29.1, 38.6]} \\
+ Participial Verb Phrase & 54.9\std{[38.0, 71.8]} & 3.3\std{[2.8, 3.8]} & {\phsz}5.8\std{[4.7, 7.0]} & 43.3\std{[41.4, 45.2]} \\
+ Participial Verb Phrase (\textit{easy}) & 88.7\std{[86.2, 91.3]} & 4.9\std{[4.3, 5.4]} & {\phsz}5.6\std{[4.6, 6.7]} & 44.5\std{[42.4, 46.5]} \\
\midrule
Transformer & 0.0\std{[0.0, 0.0]} & 0.4\std{[0.2, 0.6]} & {\phsz}8.4\std{[8.2, 8.6]} & 81.3\std{[80.7, 81.9]} \\
+ Preposing & 18.6\std{[16.2, 21.1]} & 0.5\std{[0.4, 0.6]} & {\phsz}8.6\std{[8.3, 8.9]} & 81.6\std{[81.0, 82.2]} \\
+ Preposing + Filler Words & 20.5\std{[19.1, 22.0]} & 1.3\std{[1.0, 1.6]} & {\phsz}9.1\std{[8.3, 9.8]} & 82.6\std{[82.2, 83.1]} \\
+ Participial Verb Phrase & 24.7\std{[11.8, 37.6]} & 4.2\std{[3.7, 4.7]} & 10.2\std{[9.7, 10.7]} & 83.6\std{[82.8, 84.3]} \\
+ Participial Verb Phrase (\textit{easy}) & 82.7\std{[80.9, 84.4]} & 3.8\std{[3.4, 4.2]} & {\phsz}9.9\std{[9.3, 10.5]} & 85.9\std{[85.2, 86.6]} \\
\bottomrule
\end{tabular}}
\caption{Results on the COGS structural generalization splits for different meaning preserving data augmentation methods designed to ensure that the full range of needed variable indices are seen during training. We report means (over 20 runs) with bootstrapped 95\% confidence intervals. Training on these data augmentations greatly improves compositional generalization.}
\label{tab:cogs-results-positional-indices}
\end{table*}

\begin{table*}[ht]
    \centering
    \resizebox{1.0\linewidth}{!}{
    \begin{tabular}{p{0.2\linewidth} p{0.8\linewidth}}
    \toprule
      \textbf{Variant}  & \textbf{Logical Form (LF)} \\ \midrule
      COGS & \lf{zebra(x\_1) AND need.agent(x\_2,x\_1) AND need.xcomp(x\_2,x\_4) AND walk.agent(x\_4,x\_1)} \\
      ReCOGS & \lf{zebra(47) ; need(13) AND agent(13,47) AND xcomp(13,48) AND walk(48) AND agent(48,47)} \\
      \bottomrule
    \end{tabular}}
    \caption{Semantic representations for \word{A zebra needed to walk.}\ in COGS and ReCOGS\@. LFs are detokenized for readability; tokenized examples can be found in \Tabref{tab:examples}.}    
    \label{tab:examples3}
\end{table*}

\paragraph{Preposing} One strategy to disentangle the effects of variable names and positional indices is to move the modified phrases to the front of the sentence via preposing (topicalization). This has no effect on argument structure or the meaning of the corresponding LF\@. It simply ensures that the model sees concurrences of positional indices that would otherwise have only appeared in the testing set. In our experiments, we prepose the object in 5\% of training examples containing at least one prepositional phrase.

Preposing brings a new challenge for parsing, since the model needs to learn how to parse preposed PP modifiers for object NPs as well as the regular ones. Despite this new challenge, preposing dramatically improves model performance. 

\paragraph{Filler Words} To further alleviate the problem with unseen occurrences of positional indices, we experiment with adding filler words (``um'') into the input sentence. This shifts around the variable names and positional indices in the LF (see the corresponding example in \Tabref{tab:examples}). 
We sprinkle 1--3 filler words at random in 5\% of training examples with one or more prepositional phrase. Filler words can be seen as mapping to the constant conjunct with meaning \lf{True}, and thus they do not affect the overall meaning of the LF.

\paragraph{Participial Verb Phrases} 
We augment the training set by providing examples for participial verb (PV) phrases for both objects and subjects. The original vocabulary of COGS does not have any participial verbs, so we added participial forms for all verbs.\footnote{We did this using \texttt{ChatGPT}, prompted with ``Convert the following verbs into participial verbs'' followed by a list of verbs that exist in COGS and then hand-checked the results.} To create full phrases, we randomly select a participial verb and a noun. Then, we randomly assign an article for the noun selected. For two participial verb phrases in a row, we repeat this process twice. The vocabulary files of the models are updated to include the new tokens required for parsing these PV phrases.

As shown in the corresponding row in \Tabref{tab:examples2}, the model now needs to learn this new type of noun modifier \texttt{acl} with participial verbs. We argue that testing with PP modifiers within subject NPs becomes more reasonable in this new setting, as we completely isolate the potential artifacts due to the positional indices. We add PV phrases to 15\% of training examples with maximally two PV phrases in a row. We include an \textit{easy} version without adding the novel \texttt{acl} token and replace it with the existing \texttt{nmod} token in the LF.

\paragraph{Results}
As shown in \Tabref{tab:cogs-results-positional-indices}, both the LSTM and the Transformer model start to show progress on this structural generalization split. Surprisingly, the LSTM model becomes much more effective in parsing subject NPs after including PV phrases, which contrasts with previous findings that the Transformer model is always better than the LSTM model at structural generalization on average. Our modifications isolate the effect of variable name bindings from structural generalization. The results suggest that the stagnation of model performance on this split is mostly due to the particular variable naming convention of the current semantic representation. We later show (in \Secref{sec:recogs}) that models overfit to the variable binding of COGS, which could make COGS artificially easier, and prevent us from accurately accessing compositional generalization of models.

\begin{figure*}[tp]
  \centering
  \includesvg[width=1.0\textwidth]{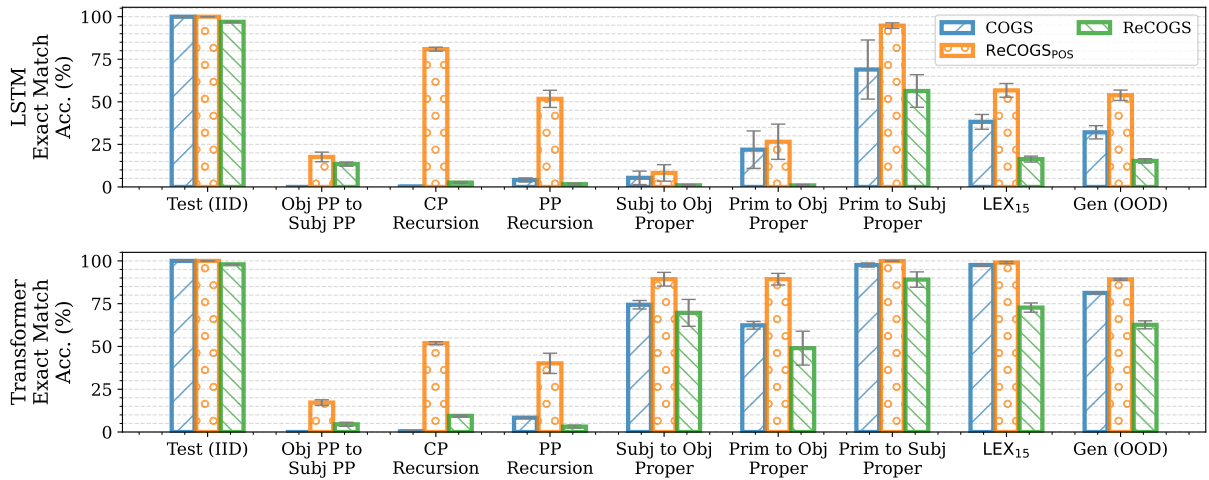}
  \caption{Model performance over different testing splits in COGS, ReCOGS$_{\text{POS}}$ (original variable name bindings are kept), and ReCOGS. We report means (of 20 runs) with bootstrapped 95\% confidence intervals.}
  \label{fig:recogs}
\end{figure*}

\section{ReCOGS: A Revised Version of COGS}\label{sec:recogs}

We now propose a revision to COGS called ReCOGS. 
We assemble the insights provided by the above experiments and use them to inform a benchmark that comes closer to assessing models purely on their ability to handle semantic generalizations.  \Tabref{tab:examples3} provides an example. We first implement the following changes, supported by our experiments, to create ReCOGS:
\begin{enumerate}\setlength{\itemsep}{0pt}
\item We remove redundant prefix tokens including \lf{x} and  \texttt{\_} (motivated by \Secref{sec:token-removal}). 
\item We replace each token position index with a random integer available in the model vocabulary file, maintaining consistent coreference. For instance, every appearance of \lf{x\_3} may be replaced with \lf{x\_46}, which makes the indices irrelevant to their token positions in the sentence. For each original COGS example, we create 5 different versions by randomly sampling 5 distinct sets of indices. We provide a ReCOGS variant, ReCOGS$_{\text{POS}}$, without this change to show its effect. 

\item We augment the current training split with longer examples by concatenating existing training examples. In total, we add in 15,360 (3,072 for each set of indices) new examples. We make sure only unique examples are kept (motivated by \Secref{sec:exp2}).

\item We prepose the object for 5\% of the training examples containing an object with at least one prepositional phrase, and we randomly add filler words (``um'') into these examples (motivated by \Secref{sec:exp3}).

\end{enumerate}
Additionally, we make the following changes that are not experimentally supported but help make ReCOGS LFs more consistent and expressive:
\begin{enumerate}\setlength{\itemsep}{0pt}
\item We treat proper nouns as predicates, which prevents collisions when multiple distinct entities share the same name (e.g., \lf{Mia(4)} and \lf{Mia(5)} can refer to different people). This also leads to a more uniform semantic treatment of noun phrases in general.
\item We prepose proper nouns and nouns with indefinite articles in the LF, parallel to definites. This makes LFs more consistent.
\item We separate situations and semantic roles, and rely on variable binding to link them (e.g., reformat \lf{agent.eat(6,7)} as \lf{eat(6) AND agent(6,7)}).
The resulting LFs more closely resemble the Neo-Davidsonian view of verbal arguments~\cite{parsons1990events}.
\end{enumerate}
\Tabref{tab:dataset-stats} provides basic dataset statistics for COGS and ReCOGS.

\begin{table}[tp]
\centering
\begin{tabular}{lrrrr}
\toprule
 &  Train & Dev & Test & Gen \\
\midrule
COGS & 24,155 & 3,000 & 3,000 & 21,000 \\
ReCOGS  & 135,547 & 3,000 & 3,000 & 21,000 \\
\bottomrule
\end{tabular}
\caption{Dataset statistics.}
\label{tab:dataset-stats}
\end{table}

We emphasize that ReCOGS continues to use variables in its representations. As we noted in \secref{sec:related}, the variable-free proposal of \citet{qiu2021improving} does not always preserve the meanings of the original COGS representations. COGS embeds variable-binding challenges that this variable-free form artificially side-steps, something that we feel is inappropriate when variable binding arguably reflects one of the deepest challenges of natural language interpretation.

As a metric for ReCOGS, we propose Semantic Exact Match (SEM)\@. For SEM, we exhaustively check whether there is a semantically consistent conversion of the variables in the predicted LFs into those of the actual LF\@. For COGS, this amounts to checking whether there is bijective map between variables in the predicted and actual LFs, where corresponding variables participate in exactly the same predications.\footnote{SEM also resembles SMATCH score~\cite{cai2013smatch} used for abstract meaning representation (AMR) parsing~\cite{banarescu-etal-2013}.} For instance, the predicted LF \lf{table(46) AND sturdy(46)} can be converted to \lf{table(1) AND sturdy(1)} via the mapping $[\lf{46} \mapsto \lf{1}]$, but the LF \lf{table(46) AND sturdy(7)} cannot be converted in this way because \lf{46} and \lf{7} would have to both map to \lf{1}. The reverse conversion is also blocked because \lf{1} would have to map to both \lf{46} and \lf{7}. In addition, we treat conjunctions in the LF as an order-free set, and exact match is evaluated as set equality after variable conversion. For the COGS language, these steps amount to checking for semantic equivalence.\footnote{Some straightforward modifications to the SEM checking procedure would be needed if COGS included  tautologies, contradictions, or equality statements between variables.}

We evaluate ReCOGS with the same two model architecture setups as in previous sections. %
\Figref{fig:recogs} reports initial results. Overall, we find that ReCOGS is a more challenging compositional generalization benchmark than COGS, but models are able to get traction on all splits, suggesting we have avoided the worrisome pattern of 0s seen in \tabref{tab:cogs-seq2seq-results}.

While both our models still achieve near 100\% performance on the IID testing split, they struggle on the structural and lexical generalization tests. However, they now show signs of life. For instance, the Object PP $\rightarrow$ Subject PP split of ReCOGS is now tractable while remaining challenging. This shows how the meaning-preserving modifications of ``um'' insertion and preposing can remove artifacts that were preventing models from learning before. In addition, our results show that LSTMs perform better than Transformers on the same split (13.4\% vs.~4.5\% on average). Future work may investigate the reason why LSTMs seem to generalize better in this scenario.

We also see the effects of breaking the relationship between variable names and linear position. Without this non-semantic pattern to rely on, models show degraded performance. To verify this, we revert ReCOGS back to the original variable name bindings of COGS where the order of LFs mirrors the word order in the input sequence, and generate a new variant, ReCOGS$_{\text{POS}}$. As shown in \Figref{fig:recogs}, performance of both models increases significantly on ReCOGS$_{\text{POS}}$. Consequently, we believe that ReCOGS poses additional challenges in terms of long-term coindexing over entities, challenges that were partly obscured by the variable naming scheme chosen for COGS.

\section{Discussion and Conclusion} 

With compositional generalization benchmarks, we hope to gain reliable information about whether models have learned to construct the \emph{meanings} of natural language sentences. However, meanings are highly abstract entities that we do not know how to specify directly, so we are compelled to conduct these assessments using LFs, which are syntactic objects that we presume can themselves be mapped to meanings in the relevant sense. The central difficulty here is that there is no single privileged class of LFs to use for this purpose, and different LFs are likely to pose substantially different learning problems and thus may lead to very different conclusions about our guiding question.

We explored these issues in the context of COGS, a prominent compositional generalization benchmark. COGS includes sub-tasks that even our best present-day models cannot get any traction on. Our central finding is that there are two major factors contributing to this abysmal performance: redundant symbols in COGS LFs and the requirement imposed by COGS that models predict the exact numerical values of bound variables. These details cannot be justified semantically, and they play a large role in shaping model performance. We showed that simple, meaning-preserving modifications to COGS fully address these problems and allow models to succeed. In turn, we propose ReCOGS, a modified version of COGS that incorporates these insights. Our models are able to get some traction on all the sub-tasks within ReCOGS, but it remains a very challenging benchmark.

One of our reviewers raised an important question relating to ReCOGS and benchmarking efforts that might follow from it: could this encourage an unproductive sort of ``LF hacking'' in which people continually reformat the data in an effort to incrementally boost performance? This is certainly a risk. However, for now, we venture that such experimentation can be productive. If the modified LFs are truth-conditionally identical to the originals, we can be confident that the generalization splits are not being compromised by this LF hacking. Thus, consistent improvements in performance may lead to lasting insights about effective meaning representation for our models, and persistent failures are likely to point to deep limitations. Overall, then, some LF hacking could help us get incrementally closer to truly evaluating the capacity of models to compute \emph{meaning}.

Our results also raise important questions about compositional generalization itself. It is in the nature of compositional generalization tasks that models will be assessed on examples that are meaningfully different from those they have seen in training. When is this fair, and when does it implicitly contradict the train set itself? Our discussion focused on the distribution of PP modifiers across different grammatical positions. COGS models are trained on examples that might suggest these are limited to object noun phrases. This may seem implausible for PP modifiers, but many natural languages do display subject/object asymmetries with regard to phenomena like case marking, definiteness, and pronominalization. In our experiments, we indirectly instructed our model about our intended generalization via meaning preserving modifications to the training data, but ultimately these issues may call for deeper changes to how we pose compositional generalization tasks.

\section*{Acknowledgements}

We would like to thank Benjamin Van Durme and anonymous reviewers for their helpful suggestions. We are also grateful to Najoung Kim, Bingzhi Li, Tal Linzen, Yuekun Yao and Linlu Qiu for very helpful discussion.

\bibliography{anthology,custom}
\bibliographystyle{acl_natbib}

\end{document}